\documentclass[letterpaper, 10 pt, conference]{ieeeconf}
\IEEEoverridecommandlockouts                              
\usepackage{cite}
\usepackage{graphicx}

\begin{document}

\newcommand{\FIG}[3]{
\begin{minipage}[b]{#1cm}
\begin{center}
\includegraphics[width=#1cm]{#2}\vspace*{-1mm}\\
{\scriptsize #3}
\vspace*{-1mm}
\end{center}
\end{minipage}
}

\newcommand{\figA}{
\begin{figure}[t]
\centering
\FIG{8}{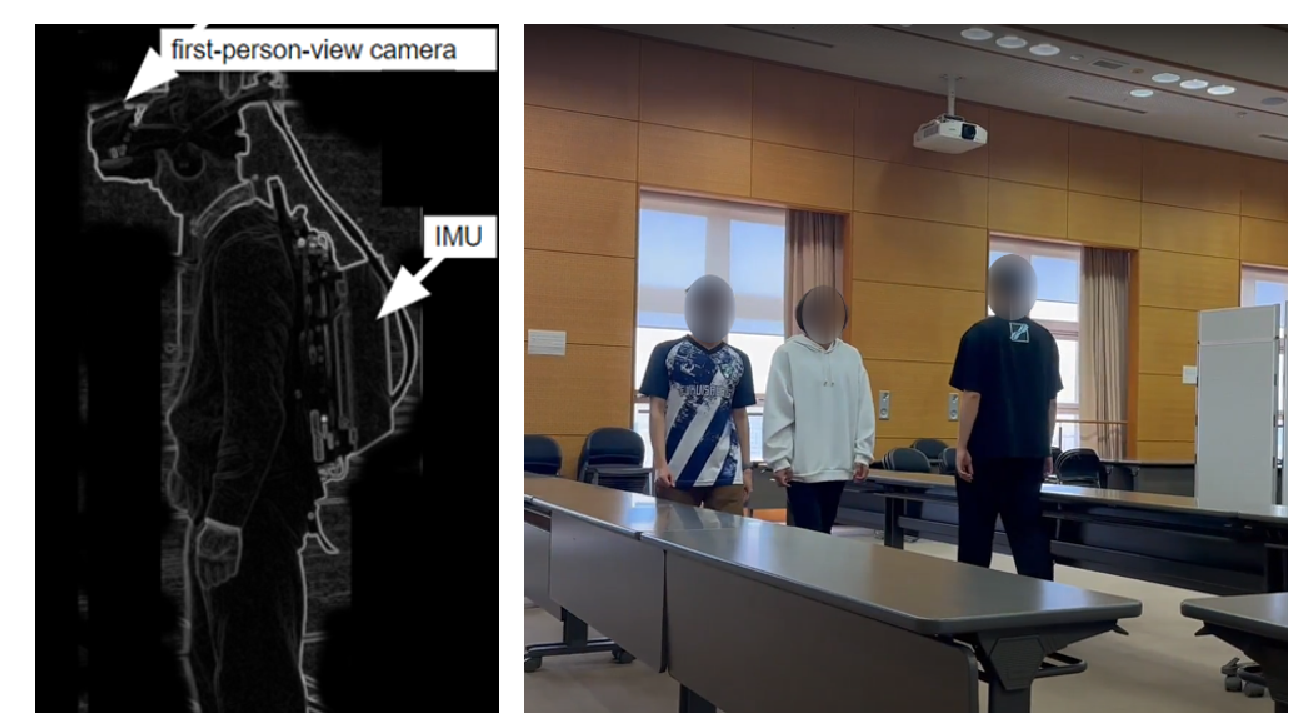}{}
\caption{Traversability prediction under severe occlusion. Left: Conventional first-person-view setup with IMU. Right: Proposed third-person-view monocular vision setup}
\label{fig:1}
\vspace*{-5mm}
\end{figure}
}

\newcommand{\figB}{
\begin{figure}[t]
\centering
\FIG{4}{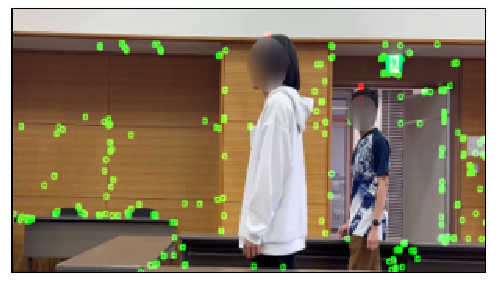}{(a)}
\FIG{4}{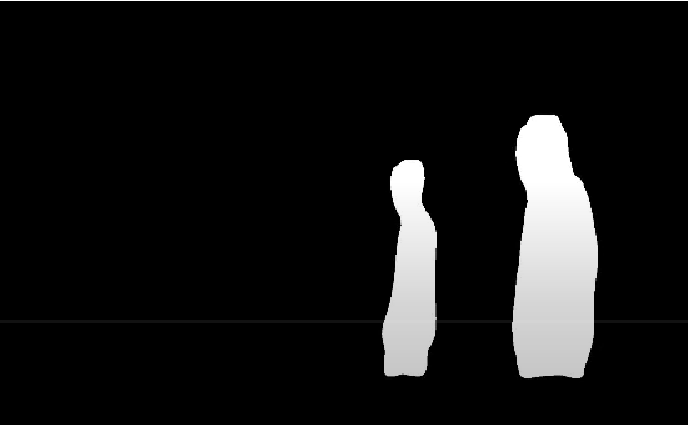}{(b)}
\FIG{4}{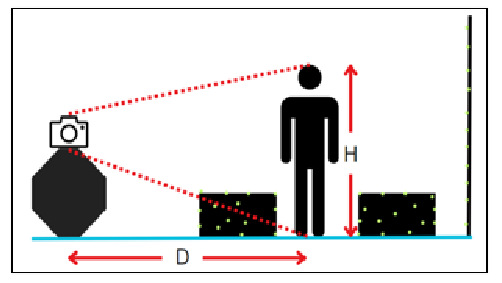}{(c)}
\FIG{4}{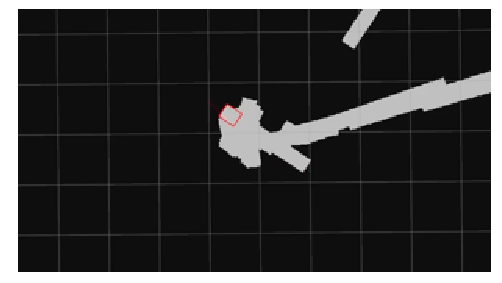}{(d)}
\caption{Scene Understanding: (a) PfH module, reference point placed on human head; (b) Human mask via Detectron2; (c) Human Height-Depth Estimation algorithm; (d) Traversability map output by PfH module. Grey = Predicted traversable region; Rectangle = human position with higher probability}\label{fig:3}
\vspace*{-5mm}
\end{figure}
}

\newcommand{\figC}{
\begin{figure}[t]
\centering
\FIG{4}{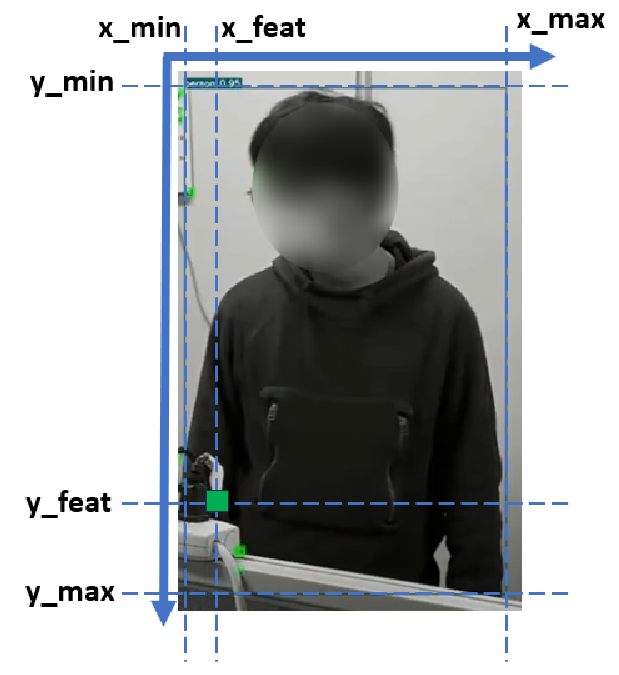}{}
\caption{Human-Object Occlusion Ordering Algorithm with parameterized occluded human region and feature points}\label{fig:4}
\vspace*{-5mm}
\end{figure}
}

\newcommand{\figD}{
\begin{figure}[t]
  \centering
  \FIG{8}{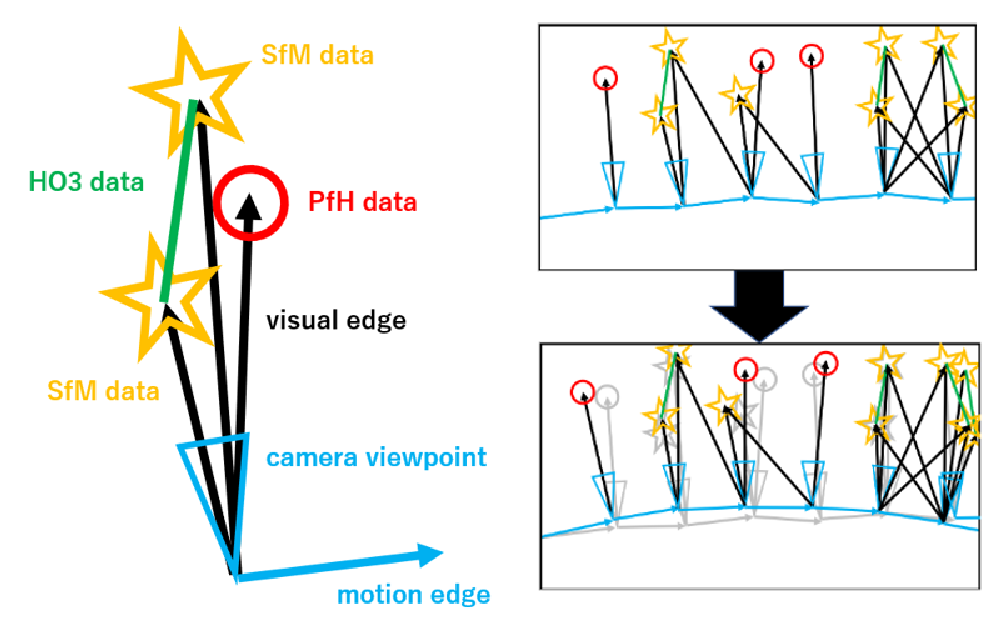}{}
  \caption{Inclusion of asynchronous map optimization events. Left: Expansion of nodes and edges in pose graph SLAM. Right: By effectively using visual edge constraints before (top) and after (bottom) asynchronous pose optimization, the optimized pose graph can be propagated to the traversability map at nearly zero cost.}\label{fig:5}
\vspace*{-5mm}
\end{figure}
}

\newcommand{\figF}{
\begin{figure}[!t]
\FIG{8}{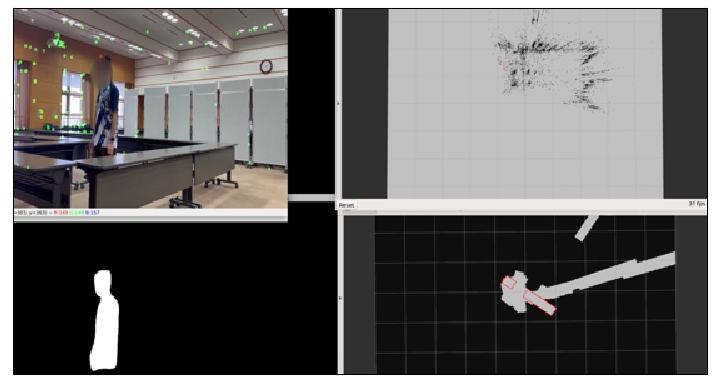}{(a)}
\FIG{4}{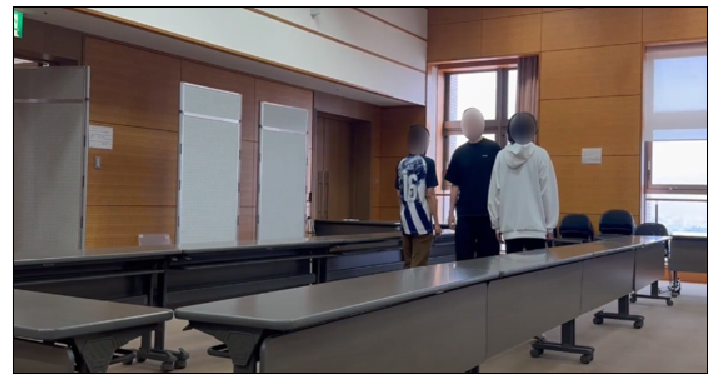}{(b)}
\FIG{4}{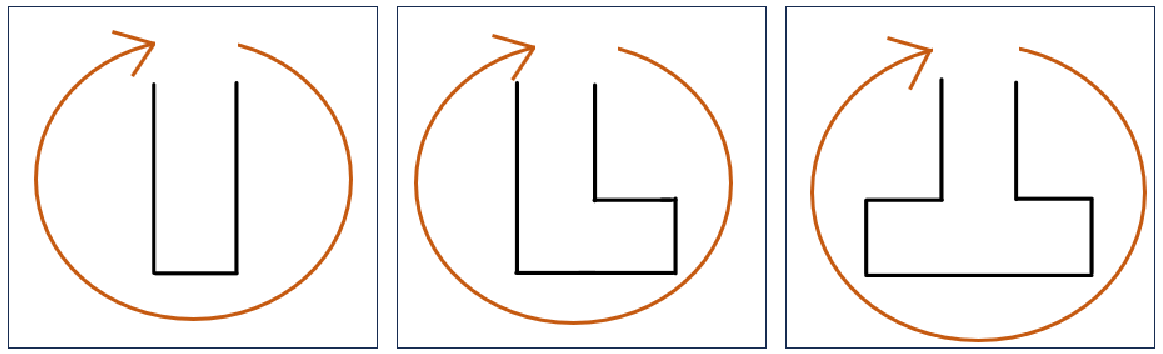}{(c)}
\caption{(a) Full framework. All the modules can be executed concurrently to obtain different traversability map and the process can be visualized. Top left: Video stream with HO3 algorithm visualized; Bottom left: Detectron2 mask; Top right: ORB-SLAM3 feature points visualized; Bottom right: Traversability map (b) I-Shape path setup. (c) Bird's eye view of obstacles setup of all kinds of configurations, the robot's path during data collection is indicated by the orange arrow}\label{fig:6}
\vspace*{-5mm}
\end{figure}
}

\newcommand{\figG}{
\begin{figure}[t!]
\FIG{4}{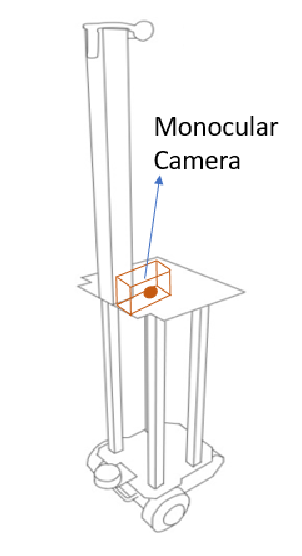}{(a)}
\FIG{4}{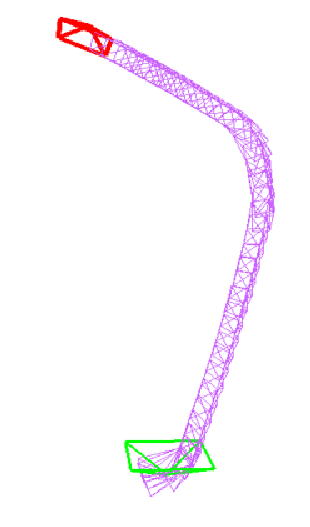}{(b)}
\caption{(a) Robot set up, with a monocular camera mounted on the platform of approx. 1m height from ground (b) Robot path taken for data collection as shown in ORB-SLAM3.}\label{fig:7}
\vspace*{-5mm}
\end{figure}
}

\newcommand{\tabA}{
\begin{table}[t]
\caption{Performance results.}
  \centering
    \begin{tabular}{l|rrr}
    \hline
& I-Cfg. & L-Cfg. & T-Cfg. \\\hline
SfM + PfH + HO3 & 2.35  & 18.68 & 15.77 \\
SfM + PfH  & 0.89  & 15.69  & 14.67 \\
SfM + HO3  & 4.53  & 16.66  & 10.53 \\
PfH + HO3  & 7.54  & 23.36  & 17.56 \\
SfM   & 3.62  &  19.04   & 29.89 \\
PfH   & 12.03 &  16.48  & 18.77 \\
HO3   & 13.38 & 10.3  & 16.45 \\\hline
\end{tabular}
  \label{tab:1}
\end{table}
}


\title{\Large \bf%
Walking = Traversable? : \\ Traversability Prediction via Multiple Human Object Tracking under Occlusion
}


\author{Jonathan Tay Yu Liang ~~~~~~~~ Kanji Tanaka
\thanks{Our work has been supported in part by JSPS KAKENHI Grant-in-Aid for Scientific Research (C) 20K12008 and 23K11270.}
\thanks{$*$J. T. Y. Liang, and K. Tanaka are with Robotics Coarse, Department of Engineering, University of Fukui, Japan. 
{\tt\small{\{mf228029@g., tnkknj@\}u-fukui.ac.jp}}
}}

\maketitle

\begin{abstract}
The emerging ``Floor plan from human trails (PfH)" technique has great potential for improving indoor robot navigation by predicting the traversability of occluded floors. This study presents an innovative approach that replaces first-person-view sensors with a third-person-view monocular camera mounted on the observer robot. This approach can gather measurements from multiple humans, expanding its range of applications. The key idea is to use two types of trackers, SLAM and MOT, to monitor stationary objects and moving humans and assess their interactions. This method achieves stable predictions of traversability even in challenging visual scenarios, such as occlusions, nonlinear perspectives, depth uncertainty, and intersections involving multiple humans. Additionally, we extend map quality metrics to apply to traversability maps, facilitating future research. We validate our proposed method through fusion and comparison with established techniques.
\end{abstract}

\IEEEpeerreviewmaketitle

\section{Introduction}

Traversability prediction, the visual recognition of whether a specific area on a 2D floor can be traversed or not, constitutes a fundamental challenge in visual robot navigation. This problem holds significant importance across various domains, including personal robotics, search and rescue operations, planetary exploration, autonomous driving, and agriculture, and has been approached through various formulations, such as obstacle detection\cite{r2}, terrain segmentation\cite{r4}, binary classification of traversability\cite{r5}, multi-class classification of traversability\cite{r6}, elevation grid mapping\cite{r7}, demonstration-based learning\cite{r8}, self-supervised learning\cite{r9}, multi-sensor fusion\cite{r10}, multi-modal fusion\cite{r11}, reinforcement learning\cite{r12}, novelty detection\cite{r13}, terrain classification\cite{r14}, and unsupervised learning\cite{r15}.

A significant issue in this field involves addressing occluded pathways, a common occurrence in indoor settings like offices, where obstacles frequently impede clear views of the floor, and finding effective solutions for this problem proves to be a formidable challenge. Current techniques struggle when significant portions of the floor are occluded, as shown in Figure \ref{fig:1}. Addressing this challenge and enhancing the ability to handle occluded regions could greatly expand the practical applications of traversability prediction.

The emerging "Floor plan from human trails (PfH)" technique in computer graphics and related fields has the potential to predict traversability in occluded floors. This technique measures human trails and interprets them as traversable regions, expanding the range of traversability prediction. Existing systems are aimed at semi-manual or semi-automatic scenarios and require first-person-view ego-motion sensors or equivalent global positioning systems. They are not suitable for third-person-view traversability prediction from an autonomous robot's vision.

\figA

This work presents a novel approach to replace first-person-view sensors with a third-person-view monocular camera equipped on the observer robot. The proposed approach can simultaneously receive measurements from multiple humans, greatly expanding its application range. The key idea is to complementary employ two types of trackers: SLAM (Simultaneous Localization and Mapping)\cite{r17} and MOT (Multi-Object Tracking)\cite{r18}. These trackers simultaneously monitor both stationary objects and moving humans and further analyze their interactions. Specifically, we frame traversability prediction as the question, ``Between which pair of stationary objects did a human pass?" and introduce a novel approach called ``Human-Object Occlusion Ordering (HO3)"\cite{r19}.

This proposed scheme serves as an add-on to off-the-shelf SLAM systems, enabling the incorporation of asynchronous map optimization events like loop closures in real time. By doing so, the proposed method achieves stable traversability prediction even in complex vision scenarios, including occlusions, non-linear perspective views, depth uncertainty, and multi-human intersections. Additionally, we extend the concept of map quality metrics to traversability map applications, aiming to support future research in this field. Traversability map is a map that defines traversability regions in the form of grid map.

We validate the effectiveness of the proposed method through comprehensive real-world experiments and by comparing it with well-known methods such as ``Structure-from-Motion (SfM)"\cite{r17} and ``(a vision-based variant of) PfH"\cite{r16}.

The contributions of this paper can be summarized as follows. (1) Exploration of traversability prediction under severe occlusions, solely relying on third-person-view monocular vision. (2) Achievement of stable prediction performance through the utilization of two types of trackers, namely SLAM and MOT, which track stationary objects and dynamic humans, enabling the observation of interactions between them. (3) Evaluation of the effectiveness of the proposed method by employing a concrete performance index for traversability maps. This evaluation is carried out through the fusion and comparison of the proposed method with the best-known methods in comprehensive real-world experiments.

\figB

\section{Approach}\label{sec:A}

Our objective is to construct a 2D grid map, specifically a traversability map (\ref{sec:A}), wherein each cell region receives a predicted traversability score. Initially, the traversability map starts with all grid cells marked as unknown, subsequently undergoing updates based on three primary methods. The first method employs SfM to track static objects, categorizing object-occupied regions as impassable (\ref{sec:B}). The second method, PfH, monitors dynamic humans, designating their paths as traversable regions (\ref{sec:C}). The third method, HO3-SLAM, analyzes interactions between static objects and dynamic humans to derive independent traversability measurements (\ref{sec:D}). These methods augment an off-the-shelf SLAM system, enabling the real-time incorporation of asynchronous map optimization events like loop closure (\ref{sec:E}). Additionally, we introduce an extended concept of map quality metrics to enhance the utility of the traversability map in various applications and support future research endeavors in this field (\ref{sec:F}).

\subsection{Transversability Map}\label{sec:A}

The traversability map is a 2D grid map applied to the robot's mobile plane, ensuring comprehensive coverage of its operational area. The grid is partitioned into cells with a spatial resolution of 10 cm $\times$ 10 cm. We found that opting for a finer cell size results in marginal performance improvements at the cost of significantly increased computational demands and storage requirements. Each grid cell can assume one of three states: ``traversable", ``untraversable" and ``unknown". During the initialization phase, every grid cell starts with an initial ``unknown" value.

\subsection{Structure-from-Motion (SfM)}\label{sec:B}

The purpose of SfM is to reconstruct objects using SLAM and interpret certain areas as ``untraversable". We utilize ORB-SLAM3\cite{r36} for SLAM, which is a cutting-edge system for real-time 3D reconstruction in unknown environments. It uses ORB feature extraction and tracking, loop closure detection, and pose graph generation and optimization. ORB-SLAM3 is known for its robustness and efficiency, making it popular in robotics and computer vision applications. Our framework relies on ORB-SLAM3 for creating a detailed 3D representation of the surroundings using a point cloud map.

\subsection{Floor Plan from Human Trails (PfH)} \label{sec:C}

The purpose of PfH is to track multiple humans even when they are hidden and interpret the trails they leave behind. However, multi-object tracking is still an area of active research with several unresolved issues. One challenge is that individual humans can enter and leave the office, and their appearance can vary significantly depending on their posture and belongings\cite{r37}. To address this, new incoming humans are assigned a unique ID\cite{r38}. Another challenge is associating appearance-based data when different humans have similar appearances, such as wearing a uniform\cite{r39}. Additionally, humans can be completely occluded by tall objects, resulting in track loss\cite{r40}. To simplify the process, we remove parts of the image sequence where the robot's turning motion is slow enough for human observation.

We use a pinhole camera model to estimate the distance between humans and the camera. 

Before using this model, we need to calibrate a key parameter ($k$) in the equation $D=kH$. $D$ represents distance measured by ORB-SLAM3, and $H$ represents the estimated height of humans (ranging from 5 to 6 feet on average). 

Once calibrated, we can transform in-image coordinates into in-map coordinates for any image feature.

\subsection{Human-Object Occlusion Ordering (HO3)}\label{sec:D}

The interaction inference question was simplified to "Between which two static objects did the human pass?" This question is based on observations of static and dynamic objects from the SLAM tracker and MOT tracker. We developed an occlusion ordering algorithm by combining human masks from Detectron2 and feature points from ORB-SLAM3. The occlusion order between humans and objects in the scene is determined by analyzing the fused information, which improves scene understanding.

The algorithm is employed to determine whether the feature points are positioned in front of or behind the human with a higher likelihood. Armed with this information, we can subsequently infer that pedestrians are likely situated amidst clusters of feature points, therefore indicating a higher probability of traversability within the intermediate area.

In our previous paper\cite{r19}, we presented an innovative approach to learning traversable areas by establishing relationships between pedestrians and object feature points. As shown in Figure\ref{fig:4}, the green point represents the feature points with coordinates (x\_feat and y\_feat), while (x\_min, x\_max, y\_min, and y\_max) represents the $x$ and $y$ range of the human area.  The current method extends the prior work\cite{r19} in several aspects: (1) Multiple human tracking (2) Moving camera tracking (3) Human distance from camera estimation. 

\figC

\subsection{Asynchrounous Map Fusion}\label{sec:E}

The information fusion module is tasked with dynamically generating the optimal traversability map, which reflects asynchronous map optimization events like SLAM's loop closure and map merging. We have developed a framework with the combination of modules SfM, PfH and HO3-SLAM.

For SfM, the traversability map can be generated on-the-fly by considering the feature points from the SLAM map as obstacles (untraversable areas). In PfH, the robot's viewpoint provides crucial information for the traversability map, with the relative position of the human observed in each frame. Each frame's view is adjusted through SLAM's asynchronous map optimization as shown in Figure \ref{fig:5}. Additionally, the traversability map is continuously updated on the fly. Regarding HO3, remember that traversability prediction involves inferring the pair of ORB features that a human passes between. Consequently, this prediction can be succinctly represented as a pair of feature IDs. This representation remains independent of the feature configuration and does not require updating during asynchronous map optimization events.

\figD

PfH and HO3-SLAM\cite{r19} modules utilize information from ORB-SLAM3\cite{r36}, specifically the coordinates of feature points and keyframes. Data transfer between the modules is facilitated through ROS nodes as shown in Figure \ref{fig:6}a.

To generate a sparse point cloud map, a video stream is input into ORB-SLAM3. Key points are detected, and their descriptors are computed, serving as potential feature points.These feature points are projected onto keyframes. The occlusion ordering algorithm is applied to the current frame using human mask detection and human detection models to determine the occlusion order of humans and feature points. The human head location is determined using the bounding box of the detection results. The PfH and HO3-SLAM modules independently predict the human's location and the traversable region. The final prediction of the traversable region is the intersection of these two areas.

\subsection{Map Quality Evaluation}\label{sec:F}

Map quality measures how well a map serves its users and varies depending on the application. In our investigation of traversability maps (SfM, PfH, HO3-SLAM), we found a gap - a lack of a suitable quality metric. To fill this gap, we propose extending map quality assessment to include traversability maps, inspired by the journey-based metric outlined in \cite{r45}. This approach considers the perspective of map users, represented by start and goal locations, and their goal to navigate using the shortest path planning algorithm. The evaluation compares each map user's shortest path with an oracle's shortest path, calculated using a manually annotated ground truth traversability map. The error is calculated as the Euclidean distance between each waypoint on the oracle's path and the nearest waypoint on the map user's path. This error is then averaged over all waypoints on the oracle's path and all map users to determine the map quality index.

\figF

\section{EXPERIMENTAL RESULTS}

\subsection{Data preparation}

We conducted an extensive data collection process to prepare three different datasets, each of which featured multiple humans and objects arranged in strategic positions within the scene to simulate crowded scenarios. To create a range of scenarios, we arranged the static objects in various configurations, including I-shaped, L-shaped, and T-shaped paths. An example of an I-shaped path setup can be seen in Figure \ref{fig:6}b. All the setups were controlled within an area of approximately 3m $\times$ 6m, and it took around 1 minute to complete a travel distance of 20m for the data collection process of each dataset.

To ensure the highest quality of data, we captured all the datasets in video format at 30 frames per second. Figure \ref{fig:6}c shows the bird's eye view of all the dataset configurations. To do this, we arranged tables of approximately 70cm $\times$ 60cm $\times$ 250cm (Height x Width x Length) to create a realistic environment for the scenarios we were simulating.

We mounted a monocular side-facing camera onto a robot and used it to capture high-quality images and videos of our experimental setup. The camera was strategically positioned on a platform of 85cm from the ground to record the environment efficiently while the robot moved around as shown in Figure \ref{fig:7}a. We input the video streams into ORB-SLAM3, a state-of-the-art algorithm for simultaneous localization and mapping, and recorded the results using the rosbag functionality to generate corresponding .bag files. These .bag files exclusively contain ORB-SLAM3 outputs with the data we collected. The robot path is as shown in Figure \ref{fig:7}b. We extract the feature points coordinate information from the ORB-SLAM3 outputs in subsequent modules to construct our map. 

To generate maps, we use the ROS map\_server\cite{r46}. This tool creates image format maps in .png for visualization or .pgm for the evaluation test. Before map evaluation and ablation studies, we need a ground truth reference. For this, we use the original ORB-SLAM3 point cloud map as a reference and manually measure and annotate objects. Note that this feature map cannot be used directly for map users' path planning, which is considered in our performance evaluation. The feature map is extended to a C-obstacle map and path planning is performed in C-space. Here, each feature point was expanded into one C-obstacle, assuming the radius of the robot to be approximately 0.5m.

\figG

To ensure thorough and accurate map evaluation, we have developed specialized code that can evaluate multiple maps simultaneously. This code assigns priorities to each map and generates an error score. By using this advanced evaluation code, we guarantee reliable and precise map evaluation, enabling us to comprehensively assess our system's performance. Additionally, this method helps us identify and address any issues or discrepancies in the maps, allowing us to make necessary adjustments for optimal system function.

\subsection{Qualitative Evaluation}

In this context, we use shorter abbreviations for the modules to represent them for better readability: SfM –  Structure from Motion, ORB-SLAM3; PfH – Floor plan from Human trails; HO3 – Human-Object Occlusion Ordering, HO3-SLAM. By employing various combinations of modules, SfM, PfH, and HO3, we can generate seven distinct combinations, namely: (SfM + PfH + HO3), (SfM + PfH), (SfM + HO3), (PfH + HO3), (SfM), (PfH) , and (HO3). Each of these configurations produces its own performance score.

To comprehensively evaluate our system, we conducted independent tests for each combination. We meticulously recorded the performance metrics and conducted thorough comparisons. 
Our experimentation encompassed three diverse datasets
(I-Cfg, L-Cfg, T-Cfg),
I-Cfg = I-Configuration of the dataset; L-Cfg = L-Configuration of the dataset; T-Cfg = T-Configuraiton of the dataset. 

For each dataset,
the performance is  evaluated using the 
map quality index in \ref{sec:F}.
The results for each method was
SfM+PfH+HO3 = (2.35, 18.68, 15.77), 
SfM+PfH  = (0.89, 15.69, 14.67),
SfM+HO3 = (4.53, 16.66, 10.53),
PfH+HO3 =  (7.54, 23.36, 17.56), 
SfM  = (3.62, 19.04, 29.89), 
PfH = (12.03, 16.48, 18.77), 
and 
HO3 = (13.38, 10.3, 16.45).
Lower is better performance.

The proposed method, (SfM + PfH + HO3), consistently performs well, particularly in simpler configurations. As the scene setup becomes more complex (L-configuration and T-configuration), the SfM (ORB-SLAM3) module does not perform as well due to sparse point cloud maps, as indicated by high error scores in the ablation studies. However, in simpler cases, the SfM module performs the best.

In terms of performance outcomes, L-configuration and T-configuration have similar results, with SfM, PfH, and HO3 modules all performing equally in both configurations. In contrast, I-configuration shows the opposite pattern. Based on these observations, the current framework is effective for simpler scenes with primarily vertical human paths. However, performance may vary when humans make turning movements around corners.

\bibliographystyle{IEEEtran}
\bibliography{ref}

\begin{thebibliography}{10}
\providecommand{\url}[1]{#1}
\csname url@rmstyle\endcsname
\providecommand{\newblock}{\relax}
\providecommand{\bibinfo}[2]{#2}
\providecommand\BIBentrySTDinterwordspacing{\spaceskip=0pt\relax}
\providecommand\BIBentryALTinterwordstretchfactor{4}
\providecommand\BIBentryALTinterwordspacing{\spaceskip=\fontdimen2\font plus
\BIBentryALTinterwordstretchfactor\fontdimen3\font minus
  \fontdimen4\font\relax}
\providecommand\BIBforeignlanguage[2]{{%
\expandafter\ifx\csname l@#1\endcsname\relax
\typeout{** WARNING: IEEEtran.bst: No hyphenation pattern has been}%
\typeout{** loaded for the language `#1'. Using the pattern for}%
\typeout{** the default language instead.}%
\else
\language=\csname l@#1\endcsname
\fi
#2}}

\bibitem{r2}
\BIBentryALTinterwordspacing
M.~Benrabah, E.~Randriamiarintsoa, C.~O. Mousse, J.~Morceaux, R.~Aufr{\`{e}}re,
  and R.~Chapuis, ``Dual occupancy and knowledge maps management for optimal
  traversability risk analysis,'' in \emph{26th International Conference on
  Information Fusion, {FUSION} 2023, Charleston, SC, USA, June 27-30,
  2023}.\hskip 1em plus 0.5em minus 0.4em\relax {IEEE}, 2023, pp. 1--6.
  [Online]. Available: \url{https://doi.org/10.23919/FUSION52260.2023.10224224}
\BIBentrySTDinterwordspacing

\bibitem{r4}
\BIBentryALTinterwordspacing
R.~O. Chavez{-}Garcia, J.~Guzzi, L.~M. Gambardella, and A.~Giusti, ``Learning
  ground traversability from simulations,'' \emph{{IEEE} Robotics Autom.
  Lett.}, vol.~3, no.~3, pp. 1695--1702, 2018. [Online]. Available:
  \url{https://doi.org/10.1109/LRA.2018.2801794}
\BIBentrySTDinterwordspacing

\bibitem{r5}
\BIBentryALTinterwordspacing
P.~Papadakis, ``Terrain traversability analysis methods for unmanned ground
  vehicles: {A} survey,'' \emph{Eng. Appl. Artif. Intell.}, vol.~26, no.~4, pp.
  1373--1385, 2013. [Online]. Available:
  \url{https://doi.org/10.1016/j.engappai.2013.01.006}
\BIBentrySTDinterwordspacing

\bibitem{r6}
\BIBentryALTinterwordspacing
M.~Wermelinger, P.~Fankhauser, R.~Diethelm, P.~Kr{\"{u}}si, R.~Siegwart, and
  M.~Hutter, ``Navigation planning for legged robots in challenging terrain,''
  in \emph{2016 {IEEE/RSJ} International Conference on Intelligent Robots and
  Systems, {IROS} 2016, Daejeon, South Korea, October 9-14, 2016}.\hskip 1em
  plus 0.5em minus 0.4em\relax {IEEE}, 2016, pp. 1184--1189. [Online].
  Available: \url{https://doi.org/10.1109/IROS.2016.7759199}
\BIBentrySTDinterwordspacing

\bibitem{r7}
\BIBentryALTinterwordspacing
Y.~Pan, X.~Xu, Y.~Wang, X.~Ding, and R.~Xiong, ``{GPU} accelerated real-time
  traversability mapping,'' in \emph{2019 {IEEE} International Conference on
  Robotics and Biomimetics, {ROBIO} 2019, Dali, China, December 6-8,
  2019}.\hskip 1em plus 0.5em minus 0.4em\relax {IEEE}, 2019, pp. 734--740.
  [Online]. Available: \url{https://doi.org/10.1109/ROBIO49542.2019.8961816}
\BIBentrySTDinterwordspacing

\bibitem{r8}
\BIBentryALTinterwordspacing
S.~Palazzo, D.~C. Guastella, L.~Cantelli, P.~Spadaro, F.~Rundo, G.~Muscato,
  D.~Giordano, and C.~Spampinato, ``Domain adaptation for outdoor robot
  traversability estimation from {RGB} data with safety-preserving loss,'' in
  \emph{{IEEE/RSJ} International Conference on Intelligent Robots and Systems,
  {IROS} 2020, Las Vegas, NV, USA, October 24, 2020 - January 24, 2021}.\hskip
  1em plus 0.5em minus 0.4em\relax {IEEE}, 2020, pp. 10\,014--10\,021.
  [Online]. Available: \url{https://doi.org/10.1109/IROS45743.2020.9341044}
\BIBentrySTDinterwordspacing

\bibitem{r9}
\BIBentryALTinterwordspacing
B.~Suger, B.~Steder, and W.~Burgard, ``Traversability analysis for mobile
  robots in outdoor environments: {A} semi-supervised learning approach based
  on 3d-lidar data,'' in \emph{{IEEE} International Conference on Robotics and
  Automation, {ICRA} 2015, Seattle, WA, USA, 26-30 May, 2015}.\hskip 1em plus
  0.5em minus 0.4em\relax {IEEE}, 2015, pp. 3941--3946. [Online]. Available:
  \url{https://doi.org/10.1109/ICRA.2015.7139749}
\BIBentrySTDinterwordspacing

\bibitem{r10}
\BIBentryALTinterwordspacing
S.~Martin, L.~Murphy, and P.~Corke, ``Building large scale traversability maps
  using vehicle experience,'' in \emph{Experimental Robotics - The 13th
  International Symposium on Experimental Robotics, {ISER} 2012, June 18-21,
  2012, Qu{\'{e}}bec City, Canada}, ser. Springer Tracts in Advanced Robotics,
  J.~P. Desai, G.~Dudek, O.~Khatib, and V.~Kumar, Eds., vol.~88.\hskip 1em plus
  0.5em minus 0.4em\relax Springer, 2012, pp. 891--905. [Online]. Available:
  \url{https://doi.org/10.1007/978-3-319-00065-7\_59}
\BIBentrySTDinterwordspacing

\bibitem{r11}
\BIBentryALTinterwordspacing
R.~Sekar, O.~Rybkin, K.~Daniilidis, P.~Abbeel, D.~Hafner, and D.~Pathak,
  ``Planning to explore via self-supervised world models,'' in
  \emph{Proceedings of the 37th International Conference on Machine Learning,
  {ICML} 2020, 13-18 July 2020, Virtual Event}, ser. Proceedings of Machine
  Learning Research, vol. 119.\hskip 1em plus 0.5em minus 0.4em\relax {PMLR},
  2020, pp. 8583--8592. [Online]. Available:
  \url{http://proceedings.mlr.press/v119/sekar20a.html}
\BIBentrySTDinterwordspacing

\bibitem{r12}
\BIBentryALTinterwordspacing
E.~F. Morales, R.~Murrieta{-}Cid, I.~Becerra, and M.~A. Esquivel{-}Basaldua,
  ``A survey on deep learning and deep reinforcement learning in robotics with
  a tutorial on deep reinforcement learning,'' \emph{Intell. Serv. Robotics},
  vol.~14, no.~5, pp. 773--805, 2021. [Online]. Available:
  \url{https://doi.org/10.1007/s11370-021-00398-z}
\BIBentrySTDinterwordspacing

\bibitem{r13}
\BIBentryALTinterwordspacing
A.~Valada, J.~Vertens, A.~Dhall, and W.~Burgard, ``Adapnet: Adaptive semantic
  segmentation in adverse environmental conditions,'' in \emph{2017 {IEEE}
  International Conference on Robotics and Automation, {ICRA} 2017, Singapore,
  Singapore, May 29 - June 3, 2017}.\hskip 1em plus 0.5em minus 0.4em\relax
  {IEEE}, 2017, pp. 4644--4651. [Online]. Available:
  \url{https://doi.org/10.1109/ICRA.2017.7989540}
\BIBentrySTDinterwordspacing

\bibitem{r14}
L.~Tai, S.~Li, and M.~Liu, ``Autonomous exploration of mobile robots through
  deep neural networks,'' \emph{International Journal of Advanced Robotic
  Systems}, vol.~14, no.~4, p. 1729881417703571, 2017.

\bibitem{r15}
\BIBentryALTinterwordspacing
R.~Hadsell, P.~Sermanet, J.~Ben, A.~Erkan, M.~Scoffier, K.~Kavukcuoglu,
  U.~Muller, and Y.~LeCun, ``Learning long-range vision for autonomous off-road
  driving,'' \emph{J. Field Robotics}, vol.~26, no.~2, pp. 120--144, 2009.
  [Online]. Available: \url{https://doi.org/10.1002/rob.20276}
\BIBentrySTDinterwordspacing

\bibitem{r17}
\BIBentryALTinterwordspacing
G.~Bresson, Z.~Alsayed, L.~Yu, and S.~Glaser, ``Simultaneous localization and
  mapping: {A} survey of current trends in autonomous driving,'' \emph{{IEEE}
  Trans. Intell. Veh.}, vol.~2, no.~3, pp. 194--220, 2017. [Online]. Available:
  \url{https://doi.org/10.1109/TIV.2017.2749181}
\BIBentrySTDinterwordspacing

\bibitem{r18}
\BIBentryALTinterwordspacing
M.~Bashar, S.~Islam, K.~K. Hussain, M.~B. Hasan, A.~B. M.~A. Rahman, and M.~H.
  Kabir, ``Multiple object tracking in recent times: {A} literature review,''
  \emph{CoRR}, vol. abs/2209.04796, 2022. [Online]. Available:
  \url{https://doi.org/10.48550/arXiv.2209.04796}
\BIBentrySTDinterwordspacing

\bibitem{r19}
J.~T.~Y. Liang and T.~Kanji, ``Ho3-slam: Human-object occlusion ordering as
  add-on for enhancing traversability prediction in dynamic slam,'' in
  \emph{2023 IEEE 13th International Conference on Pattern Recognition Systems
  (ICPRS)}.\hskip 1em plus 0.5em minus 0.4em\relax IEEE, 2023, pp. 1--7.

\bibitem{r16}
\BIBentryALTinterwordspacing
C.~Mura, R.~Pajarola, K.~Schindler, and N.~J. Mitra, ``Walk2map: Extracting
  floor plans from indoor walk trajectories,'' \emph{Comput. Graph. Forum},
  vol.~40, no.~2, pp. 375--388, 2021. [Online]. Available:
  \url{https://doi.org/10.1111/cgf.142640}
\BIBentrySTDinterwordspacing

\bibitem{r36}
\BIBentryALTinterwordspacing
C.~Campos, R.~Elvira, J.~J.~G. Rodr{\'{\i}}guez, J.~M.~M. Montiel, and J.~D.
  Tard{\'{o}}s, ``{ORB-SLAM3:} an accurate open-source library for visual,
  visual-inertial, and multimap {SLAM},'' \emph{{IEEE} Trans. Robotics},
  vol.~37, no.~6, pp. 1874--1890, 2021. [Online]. Available:
  \url{https://doi.org/10.1109/TRO.2021.3075644}
\BIBentrySTDinterwordspacing

\bibitem{r37}
\BIBentryALTinterwordspacing
H.~Zhang, C.~M. Reardon, and L.~E. Parker, ``Real-time multiple human
  perception with color-depth cameras on a mobile robot,'' \emph{{IEEE} Trans.
  Cybern.}, vol.~43, no.~5, pp. 1429--1441, 2013. [Online]. Available:
  \url{https://doi.org/10.1109/TCYB.2013.2275291}
\BIBentrySTDinterwordspacing

\bibitem{r38}
\BIBentryALTinterwordspacing
D.~Wu, S.~Zheng, X.~S. Zhang, C.~Yuan, F.~Cheng, Y.~Zhao, Y.~Lin, Z.~Zhao,
  Y.~Jiang, and D.~Huang, ``Deep learning-based methods for person
  re-identification: {A} comprehensive review,'' \emph{Neurocomputing}, vol.
  337, pp. 354--371, 2019. [Online]. Available:
  \url{https://doi.org/10.1016/j.neucom.2019.01.079}
\BIBentrySTDinterwordspacing

\bibitem{r39}
\BIBentryALTinterwordspacing
D.~Kuplyakov, E.~Shalnov, and A.~Konushin, ``Markov chain monte carlo based
  video tracking algorithm,'' \emph{Program. Comput. Softw.}, vol.~43, no.~4,
  pp. 224--229, 2017. [Online]. Available:
  \url{https://doi.org/10.1134/S0361768817040053}
\BIBentrySTDinterwordspacing

\bibitem{r40}
\BIBentryALTinterwordspacing
J.~Zhu, H.~Yang, N.~Liu, M.~Kim, W.~Zhang, and M.~Yang, ``Online multi-object
  tracking with dual matching attention networks,'' \emph{CoRR}, vol.
  abs/1902.00749, 2019. [Online]. Available:
  \url{http://arxiv.org/abs/1902.00749}
\BIBentrySTDinterwordspacing

\bibitem{r45}
\BIBentryALTinterwordspacing
A.~M. Andrew, ``\emph{The Map-Building and Exploration Strategies of a Simple
  Sonar-Equipped Mobile Robot: An Experimental Quantitative Evaluation}, by
  david lee, distinguished dissertations in computer science series, cambridge
  university press, cambridge, 1996, xi+228 pp., {ISBN} 0-521-57331-9 (hbk:
  {\textsterling}35),'' \emph{Robotica}, vol.~15, no.~2, pp. 233--236, 1997.
  [Online]. Available: \url{https://doi.org/10.1017/s0263574797240257}
\BIBentrySTDinterwordspacing

\bibitem{r46}
\BIBentryALTinterwordspacing
{Stanford Artificial Intelligence Laboratory et al.}, ``Robotic operating
  system.'' [Online]. Available: \url{https://www.ros.org}
\BIBentrySTDinterwordspacing

\end{thebibliography}

\end{document}